\documentclass[runningheads]{llncs}
\usepackage[T1]{fontenc}
\usepackage{graphicx}
\usepackage{multirow}
\usepackage{array}
\usepackage{booktabs}
\usepackage{tabularray}
\usepackage{algorithm}
\usepackage{algpseudocode}
\usepackage{graphicx,verbatim}
\usepackage{amsmath}  
\usepackage{amssymb}   

\usepackage{xcolor} 
\usepackage{mathtools} 
\begin{document}
\newcommand{\equalcontrib}{\textsuperscript{\textdagger}}
\title{GRASP-PsONet: Gradient-based Removal of Spurious Patterns for PsOriasis Severity Classification}
\titlerunning{GRASP-PsONet}
%
\author{Basudha Pal\inst{1,2}\equalcontrib\and 
Sharif Amit Kamran\inst{1}\and
Brendon Lutnick\inst{1}\and
Molly Lucas\inst{1}\and
Chaitanya Parmar\inst{1}\and
Asha Patel Shah\inst{1}\and
David Apfel\inst{1}\and
Steven Fakharzadeh\inst{1}\and
Lloyd Miller\inst{1}\and
Gabriela Cula\inst{1}\and
Kristopher Standish\inst{1}}
%
\authorrunning{B. Pal et al.}
%
\institute{Johnson\&Johnson Innovative Medicine, New Brunswick, NJ, USA \and
Johns Hopkins University, Baltimore, MD, USA \\
\email{bpal5@jhu.edu}}

%
\maketitle              
\begin{abstract}
Psoriasis (PsO) severity scoring is vital for clinical trials but is hindered by inter-rater variability and the burden of in-person clinical evaluation. Remote imaging utilizing patient-captured mobile photos offers scalability but introduces challenges, such as variations in lighting, background, and device quality that are often imperceptible to humans but may impact model performance. These factors, coupled with inconsistencies in dermatologist annotations, reduce the reliability of automated severity scoring. We propose a framework to automatically flag problematic training images that introduce biases and reinforce spurious correlations which degrade model generalization by using a gradient-based interpretability approach. By tracing the gradients of misclassified validation images, we detect training samples where model errors align with inconsistently rated examples or are affected by subtle, non-clinical artifacts. We apply this method to a ConvNeXT-based weakly supervised model designed to classify PsO severity from phone images. Removing 8.2\% of flagged images improves model AUC-ROC by 5\% (85\% to 90\%) on a held-out test set. Commonly, multiple annotators and an adjudication process ensure annotation accuracy, which is expensive and time-consuming. Our method correctly detects training images with annotation inconsistencies, potentially eliminating the need for manual reviews. When applied to a subset of training images rated by two dermatologists, the method accurately identifies over 90\% of cases with inter-rater disagreement by rank-ordering and reviewing only the top 30\% of training data. This framework improves automated scoring for remote assessments, ensuring robustness and scalability despite variability in data collection. Our method handles both inconsistencies in image conditions and annotations, making it ideal for applications lacking standardization of controlled clinical environments.
\end{abstract}
\let\thefootnote\relax\footnotetext{\equalcontrib This work was done during an internship at J\&J Innovative Medicine}
\section{Introduction}
Psoriasis (PsO) is a chronic systemic inflammatory disease that affects 2\%-3\% of the global population and is associated with comorbidities such as psoriatic arthritis (PsA), diabetes, and cardiovascular diseases \cite{zhou2015dermatology,berth2006study}. PsO severity assessment is essential for decision-making during clinical trials, as treatment selection is based on disease severity. 
Dermatologists use the Psoriasis Area and Severity Index (PASI) as the gold standard for quantifying PsO, scoring lesion extent and severity on a scale of 0–72 \cite{berth2006study}. Cases are then classified as mild, moderate, or severe based on predefined thresholds. However, in-clinic evaluations impose a logistical burden on both patients and physicians.  Combining remote imaging, where patients capture images using mobile devices, with deep learning-based automated PASI scoring offers a scalable alternative. This approach reduces the need for in-person visits, minimizes subjectivity, and streamlines disease monitoring \cite{li2020psenet,huang2023artificial,schaap2022image,kamran2025pso}. Despite advancements in deep learning-based PsO assessment, existing models face several limitations, such as extensive pre-processing using bounding boxes \cite{li2020psenet}, background removal \cite{li2020psenet,huang2023artificial,schaap2022image}, and exclusion of clinically relevant regions \cite{xing2024deep,schaap2022image} which are difficult to detect. Some approaches use separate models for regional scoring \cite{kamran2025pso} or distinct computations for erythema, induration, desquamation, and lesion area ratio \cite{pal2016severity,pal2018severity} adding to complexity.\\
\indent Despite being scalable, remote imaging can introduce significant variability in illumination, background, and device quality, which can lead to spurious correlations and degrade model performance. Furthermore, annotation inconsistencies among dermatologists contribute to unreliable training data, making it difficult for models to generalize \cite{resneck2004dermatology,liu2020deep,yao2018study}. A promising direction for addressing these challenges is data attribution, which aims to identify training samples that negatively impact model generalization. Koh \textit{et al.} \cite{koh2017understanding} and Yeh \textit{et al.} \cite{yeh2018representer} extended the concept of influence functions from robust statistics to deep learning. Researchers have explored feature-level influence by estimating how specific input features impact individual predictions \cite{sundararajan2017axiomatic,lundberg2017unified,ribeiro2016should}, as well as training sample influence by assessing how data points contribute to overall model performance \cite{owen2017shapley,owen2014sobol}. Gradient-tracing methods, such as TracIn \cite{pruthi2020estimating} and its practical adaptation TracInCP, estimate training sample influence by tracking gradient updates across minibatches and checkpoints. While \cite{hara2019data}, a closely related work, leverages Hessians to measure influence in the absence of a sample, TracInCP replays training via stored checkpoints to approximate its effect on test predictions. To emphasize on model explainability in medical imaging, some researchers have leveraged influence functions to analyze and interpret decision-making \cite{hossain2025explainable}. Recent work in medical imaging has leveraged model pruning to improve performance and reduce computation \cite{bayasi2021culprit,holste2023does}. The Dynamic Average Dice score by He \textit{et al.} \cite{he2023data} on the other hand, focuses on data pruning which dynamically quantifies the importance of each training sample by assessing its contribution to the Dice coefficient, allowing a score guided identification and removal of non-informative training samples. To the best of our knowledge, such score-based data pruning methods have been less explored in medical imaging.\\
\indent In this study, we introduce GRASP-PsONet, a gradient-tracing based influence estimation method for efficient data pruning in PsO severity classification. Given the challenge that a training image may be detrimental to one validation instance yet beneficial to another, GRASP-PsONet optimizes data selection. Our key contributions are as follows:

\begin{itemize}
	    \item{GRASP-PsONet is built on an existing weakly supervised multi-instance learning (MIL) framework, provides an end-to-end solution which eliminates extensive pre-processing (e.g., bounding boxes, region-specific models) while remaining resilient to spurious correlations and annotation inconsistencies. Using gradient tracing, our method enhances model interpretability and generalizability by identifying and removing influential training samples that contribute to misclassifications in the validation set. Specifically, we compute influence scores from misclassified validation examples and systematically prune \(2.8\%\)-\(13.3\%\) of the training dataset, improving overall robustness.}
	
	\item{We use self-influence scores to identify potential mislabeled examples \cite{pruthi2020estimating}. Self-influence quantifies how removing a training example affects its own prediction, with high scores indicating mislabeled or atypical samples. We rank-order training images based on self-influence and demonstrate that this method effectively detects annotation inconsistencies. When applied to a subset rated by two dermatologists, reviewing only the top 30\% of ranked data correctly identified 90.3\% of inter-observer disagreements.}
	
	\item{Using data attribution based method boosts performance on the multi-class PsO classification task by improving AUC for both readers by approximately 5\% and 10\% for ConvNeXT and ViT-based encoders, respectively.}
\end{itemize}

\section{Methodology}

\subsection{Problem Setup}
The data was divided into a training set, \( S_{\text{train}} = \{(x_1^{\text{trainpv}}, y_1^{\text{trainpv}}), \dots, \\(x_N^{\text{trainpv}}, y_N^{\text{trainpv}})\} \), with \( N = 46 \) images per patient visit (pv) and 610 unique patient visits. The validation set was \( S_{\text{val}} = \{(x_1^{\text{valpv}}, y_1^{\text{valpv}}), \dots, (x_N^{\text{valpv}}, y_N^{\text{valpv}})\} \), where each label \( y \) denotes PsO severity: mild (\(\text{PASI}: 0-5\)), moderate (\(\text{PASI}: 5-10\)), or severe (\(\text{PASI} > 10\)) based on all 46 images per visit. 
Influence is the reduction in loss on a validation example \( z' \in S_{\text{val}} \) caused by using a training example \( z \in S_{\text{train}} \). The aim is to find influential training images that lead to validation misclassifications in an existing MIL model and retrain it after removing these images. The final model is chosen based on the best validation AUC. Training labels come from Reader 1, while evaluation uses scores from two readers. For each misclassified validation visit, the image with the highest attention score among the 46 is chosen, and gradients are traced to remove the top \( k \) influential training images, \( x_{\text{rem}} \in S_{\text{train}} \). After removing influential images, \( N \) becomes variable as it decreases with the number of removed images. `Baseline' refers to the MIL models without data attribution for both ConvNeXT and ViT encoders.
\begin{figure}[t]
	\centering
	\includegraphics[width=0.9\linewidth]{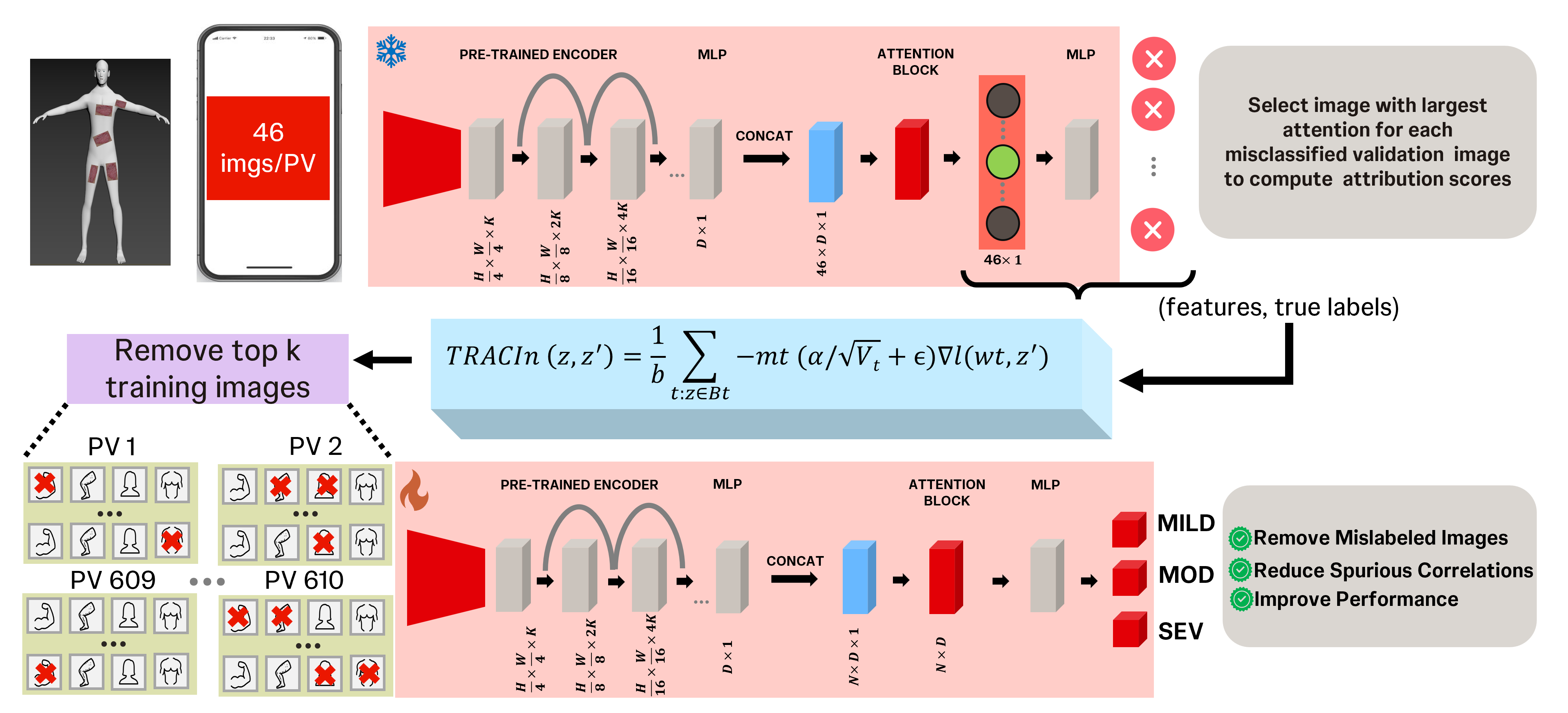}
	\caption{The overall MIL framework for multiclass PsO severity classification, including data attribution.}
	\label{fig:mil}
\end{figure}

\subsection{Overall Architecture}
Our architecture consists of a pre-trained encoder, an attention block, and a multi-layer perceptron (MLP) for multi-class classification, as shown in Fig.\ref{fig:mil}. Given that our dataset includes body images captured by patients, we found ImageNet \cite{deng2009imagenet} pre-trained encoders effective for transfer learning. We evaluated encoder architectures such as ConvNeXT and Vision Transformer (ViT). Here we describe the pre-trained encoder for multiclass PsO classification using ConvNeXT as we obtain best Baseline results using this encoder. The ConvNeXT encoder employs convolutional and downsampling blocks, producing feature dimensions of \(R^{H/4 \times W/4 \times K}\), \(R^{H/8 \times W/8 \times 2K}\), \(R^{H/16 \times W/16 \times 4K}\), and \(R^{H/32 \times W/32 \times 8K}\) for an input RGB image of size \(H \times W \times C\) where K is the number of channels. The encoder output feature dimension is reshaped to \(R^{D \times 1}\) via an MLP where \(D=768\). This is repeated \(N\) times, where \(N\) corresponds to the images per pv (initially 46/pv). The resulting features are concatenated into a tensor of size \(R^{N \times D \times 1}\), input into the attention block, and the output is passed to a final MLP to generate class probabilities for three classes. 


\begin{algorithm}
	\caption{\textbf{: Tracing Gradients in \textsc{GRASP}-PsONet for TracIn Score with Adam}}
	\label{alg1}
	\begin{algorithmic}[1]
		\State \textbf{Input:} Model $F$, validation point $z' \in S_{\text{val}}$, training point $z \in S_{\text{train}}$, training checkpoints $\{w_t\}_{t=1}^{T}$, mini-batches $B_t$ for $t = 1,2,\dots,T$, batch size $b$, Adam moments $m_t, v_t$, learning rate $\eta_t$, loss function $\mathcal{L}(w,z)$.
		\State \textbf{Output:} TracIn attribution score $\text{TracIn}(z',z)$.
		\State \textbf{Initialize:} $\text{TracIn}(z',z) \gets 0$
		\For{$t = 1$ \textbf{to} $T$}
		\If{$z \in B_t$}
		\State Compute gradients: $\nabla_{\theta} \mathcal{L}(w_t, z')$ (w.r.t. $z'$) and $\nabla_{\theta} \mathcal{L}(w_t, z)$ (w.r.t. $z$)
		\State Update the TracIn score:
		\[
		\text{TracIn}(z',z) \gets \text{TracIn}(z',z) + \eta_t \cdot \frac{m_t}{\sqrt{v_t} + \epsilon} \cdot \nabla_{\theta} \mathcal{L}(w_t, z') \cdot \nabla_{\theta} \mathcal{L}(w_t, z)
		\]
		\EndIf
		\EndFor
		\State Normalize by batch size: $\text{TracIn}(z',z) \gets \frac{1}{b} \text{TracIn}(z',z)$
		\State \textbf{return} $\text{TracIn}(z',z)$
	\end{algorithmic}
\end{algorithm}
\subsection{Removing Influential Images}  
We compute the impact of individual training samples on validation predictions by integrating gradient-based data attribution \cite{pruthi2020estimating} as outlined in Algorithm \ref{alg1}, which estimates influence scores by computing gradient similarity between training and validation examples.  For each misclassified validation image, we rank training samples based on their cumulative influence and remove the \( k \) most influential ones, refining \( S_{\text{train}} \). In the MIL model, we handle this by passing a binary vector indicating image presence, which is multiplied with the attention block to mask absent images.
 The influence of a training sample \( z \) on a validation example \( z' \) is quantified by the TracIn score, \( \text{TracIn}(z', z) \), which approximates the loss reduction in \( z' \) when \( z \) is utilized during training. Specifically, \( \mathcal{L}(w_{t+1}, z') = \mathcal{L}(w_t, z') + \nabla \mathcal{L}(w_t, z') \cdot (w_{t+1} - w_t) + \mathcal{O}(\|w_{t+1} - w_t\|^2) \). Ignoring higher-order terms yields \( \mathcal{L}(w_{t+1}, z') = \mathcal{L}(w_t, z') + \nabla \mathcal{L}(w_t, z') \cdot (w_{t+1} - w_t) \). For Adam optimizers, \( w_{t+1} - w_t = -\eta_t \frac{m_t}{\sqrt{v_t} + \epsilon} \), where \( m_t \) and \( v_t \) are the first and second moments, \( \eta_t \) is the learning rate, and \( \epsilon \) ensures numerical stability. Substituting, we get \( \mathcal{L}(w_{t+1}, z') - \mathcal{L}(w_t, z') = \text{TracIn}(z', z) = \eta_t \frac{m_t}{\sqrt{v_t} + \epsilon} \nabla \mathcal{L}(w_t, z') \). For mini-batches \( B_t \) with batch size \( b \geq 1 \), the final computation of \(\text{TracIn}(z', z)\) is shown in Algorithm~\ref{alg1}. This method identifies training images that drive misclassifications due to spurious correlations like lighting or labeling errors. Since our labels are at the patient visit level, with each visit containing 46 images, we track loss evolution on the most attended image per visit for misclassified validation patients to find influential training examples. We are able to maintain data versatility as we get a score for each image and selectively drop images rather than removing all 46 images of a particular patient.

\section{Experiments}
\subsection{Dataset and Settings}

The dataset consisted of 344 screened patients (220 female, 124 male) who each had 1–4 unique visits as part of the study protocol (baseline, weeks 2, 4, and 8), resulting in 844 total visits. Each visit comprised 46 images, leading to a dataset of 38,824 total images. Data were split into training (70\%; 610 visits, 247 patients), validation (10\%; 64 visits, 28 patients), and test (20\%; 170 visits, 69 patients) sets. Skin tones were categorized using Fitzpatrick types: I \((N=60)\), II \((N=163)\), III \((N=85)\), IV \((N=22)\), V \((N=12)\), and VI \((N=2)\). PASI scores were assigned by one of seven dermatologists from a contracted research organization (Reader 1), with inter-rater variability assessed by an independent eighth rater (Reader 2).

\subsection{Implementation Details}
Our models were trained using PyTorch \cite{paszke2019pytorch}. Images were resized to \(224 \times 224\) and normalized with ImageNet statistics: mean \((0.485, 0.456, 0.406)\) and std \((0.229, 0.224, 0.225)\). Training used the Adam optimizer \cite{kingma2014adam} with \(\alpha = 10^{-6}\), weight decay \(10^{-4}\), batch size 4, and 100 epochs on four NVIDIA A100 GPUs. Weighted sampling addressed patient-level imbalanced PASI distributions.

\subsection{Downstream Task and Evaluation Metrics}
We implement a gradient tracking data attribution algorithm using pretrained PsO classification checkpoints for the Baseline model and misclassified validation examples. We identify the top-k influential training points causing misclassifications by computing influence scores using labels from Reader 1 and retrain our MIL model after removing these k points, comparing its performance to the Baseline. We control the number of removed images to avoid negative effects on training. In our analysis, 9 out of 64 validation patient visits (each with 46 images) are misclassified. For each misclassified visit, we select the image with the highest attention score to identify the top-100, 200, 300, 400, and 500 influential training samples, representing 3-16\% of the training dataset (900–4,500 images out of 28,060). We observe an overlap in flagged training images across misclassified cases, leading to fewer removals than the calculated maximum. For example, targeting the top-500 images for each misclassified validation image could remove up to 4,500 images, but due to overlap, we remove only 3,734 unique images. Thus, the actual removal ranges from 2.8\% to 13.3\% of the training set. This controlled removal keeps the dataset large enough for effective training while reducing harmful samples. We evaluate this on a multiclass PsO severity classification task, finding that removing influential points and retraining improves AUC-ROC and Cohen's Kappa across two readers. 
\subsection{Detecting Annotation Inconsistencies}
Currently, we have 342 patient visits scored by two readers comprising the entire test set (170 patient visits) and 172 visits from the train set. To assess our algorithm's ability to detect annotation inconsistencies, we conduct an experiment by reconstructing our dataset to include the test data that were independently scored by two readers making it have 780 patient visits. This is possible as we do not want to evaluate further, rather just analyze if our method is able to flag images with annotation inconsistencies. Following prior work \cite{pruthi2020estimating}, we utilize self-influence scores, which quantify a training sample's influence on its own loss. As computing self-influence scores is computationally heavy, out of these 342 doubly rated patient visits, we pick 100 patient visits at random but maintain the ratio of same label:different label to approximately 84:16 which is the same as that of the doubly rated dataset. For each of these 100 patient visits, we compute a 46 $\times$ 46 self-influence matrix and assign the maximum diagonal value as the self-influence score for that visit. Higher self-influence scores are expected to indicate potential mislabeling, enabling systematic identification of annotation discrepancies.
	

\begin{table}[t]
	
	\centering
	
	\caption{Performance comparison of ConvNeXT and ViT on the test set using Cohen's Kappa and AUC with different numbers of images removed from the training set.}
	
	\label{tab:results}
	
	\renewcommand{\arraystretch}{1.1} 
	
	\setlength{\tabcolsep}{3pt} 

	\resizebox{\textwidth}{!}{ 
		
		\begin{tabular}{l|c|c|cc|cc}  
			
			\toprule
			
			\textbf{Encoder} & \textbf{k-value} & \textbf{\% Training Images Removed } & \multicolumn{2}{c|}{\textbf{Reader 1}} & \multicolumn{2}{c}{\textbf{Reader 2}} \\
			
			&  &  & Cohen's Kappa & AUC & Cohen's Kappa & AUC \\
			
			\midrule
			
			\multirow{6}{*}{ConvNeXT}
			
			& Baseline & 0.0\% (0/28,060)& 0.53  & 0.85 & 0.56 & 0.86 \\
			
			& Top 100 removed & 2.8\% (794/28,060)& 0.52 & 0.88 & 0.45 & 0.86 \\
			
			& Top 200 removed & 5.6\% (1,576/28,060)& 0.54 & 0.88 & 0.45 & 0.87 \\
			
			& Top 300 removed & 8.2\% (2,292/28,060)& \textbf{0.62} & \textbf{0.90} & \textbf{0.54} & \textbf{0.89} \\
			
			& Top 400 removed & 10.8\% (3,019/28,060) & 0.49 & 0.85 & 0.46 & 0.85 \\
			
			& Top 500 removed & 13.3\% (3,734/28060) & 0.42 & 0.83 & 0.38 & 0.80 \\
			
			\midrule
			
			\multirow{4}{*}{ViT}
			
			& Baseline & 0.0\% (0/28,060)& 0.37 & 0.72 & 0.39 & 0.72 \\
			
			& Top 100 removed & 2.8\% (794/28,060)& 0.41 & 0.75 & 0.39 & 0.75 \\
			
			& Top 300 removed & 8.2\% (2,292/28,060) & 0.44 & 0.81 & 0.43 & 0.78 \\
			
			& Top 500 removed & 13.3\% (3,734/28,060) & \textbf{0.61} & \textbf{0.89} & \textbf{0.52} & \textbf{0.84} \\
			
			\bottomrule
			
		\end{tabular}
		
	}
	
\end{table}

\section{Results}
\begin{figure}[h]
	\centering
	\includegraphics[width = 0.7\linewidth]{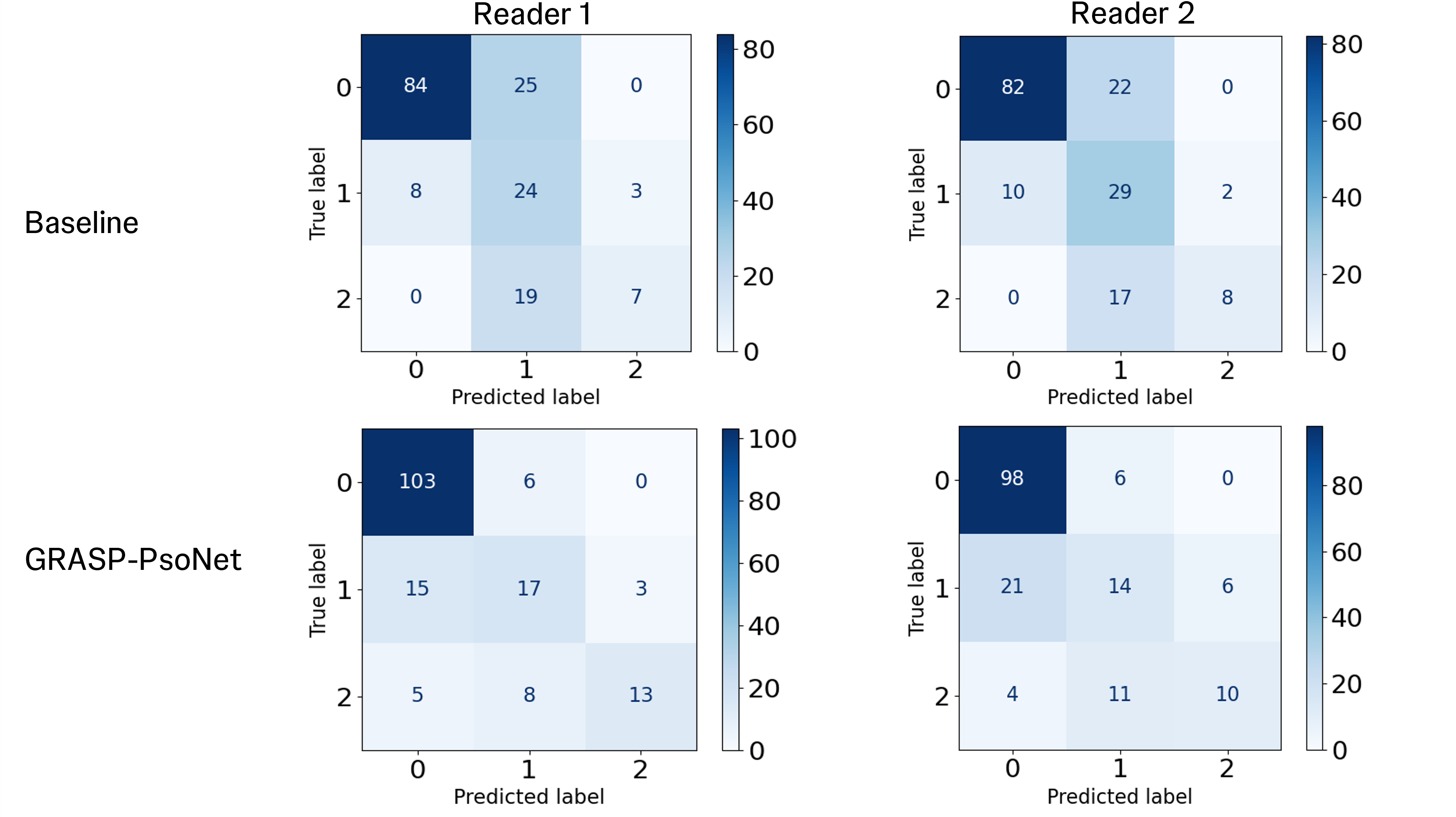}
	\caption{Confusion matrices for the Baseline and our method on the test set after removing 300 training images per validation misclassification using a ConvNeXT encoder.}
	\label{fig:cfmat}
\end{figure}

In this section, we present the quantitative results for the PsO severity classification task, and highlight the efficacy of our approach in addressing poor annotations. We report the values of micro-average AUC and linearly weighted Cohen's Kappa after removing images from the training data as summarized in \ref{tab:results}. The most favorable outcomes were obtained by removing the top 300 images from the training set per validation misclassification for multi-class severity classification using ConvNeXT encoders. We show a few additional experiments on the ViT backbone where removing top 500 images performs the best. Fig \ref{fig:cfmat} displays the confusion matrices for two independent raters on a held-out test set, comparing baseline performance with GRASP-PsONet after removing the top 300 images from training per validation misclassified patient using a ConvNeXT based encoder. This approach significantly improves performance, achieving AUC-ROC scores of 88.8\% and 90.2\% for the two raters. 
\begin{figure}[h] 
	\centering
	\includegraphics[width=0.78\linewidth]{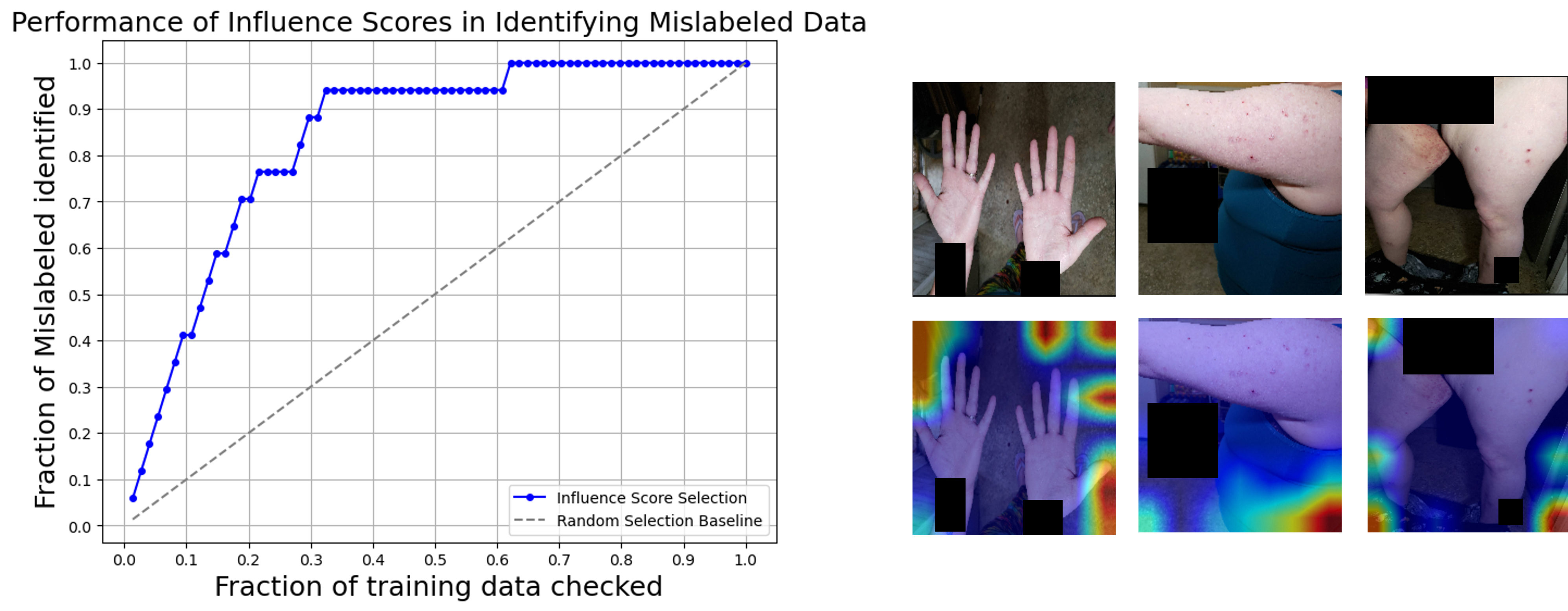}  
	\caption{\textbf{Left:} Proportion of correctly identified mislabeled samples using a rank-ordered list of self-influence scores. \textbf{Right:} Example of Grad-CAM maps on training images with inter-rater discrepancies where the model misclassified the label.}
	\label{fig:proportion}
\end{figure}

A key aspect of our study involves leveraging self-influence scores to correctly identify inter-rater discrepancies within the training dataset. Fig \ref{fig:proportion} presents a performance curve illustrating the effectiveness of this approach. The x-axis represents the fraction of the training dataset inspected, while the y-axis indicates the cumulative proportion of mislabeled samples identified. The solid blue line demonstrates that by reviewing only the top 30\% of ranked samples, we successfully identify over 90\% of the mislabeled cases. This result highlights the advantage of self-influence scores in prioritizing label verification efforts, offering an automated strategy for dataset quality checks. We also show heatmaps on some flagged training images. Interestingly, for high-influence images, GRAD-CAM reveals that the model is focusing on incorrect/irrelevant regions, suggesting that these images are problematic and affect training due to spurious correlations.

\section{Conclusion and Future Work}
In this work, we introduced a novel framework for PsO severity classification that leverages score-based influence functions to refine training data. By tracing gradients from the optimization process, we identify and remove the most influential training images using misclassified images from validation data to improve performance and generalizability. We also demonstrate the capacity of this method to specifically flag images which have discrepancies in annotations. In the future, instead of removing such problematic images, we can send them for re-scoring or quality check, thereby alleviating the need for complete dataset checking and re-scoring. 

\begin{credits}
\noindent\textbf{Acknowledgments.} This work was sponsored by Johnson\&Johnson Innovative Medicine. This version of the contribution has been accepted for publication after peer review but is not the Version of Record and does not reflect post-acceptance improvements or any corrections.
\end{credits}

\bibliographystyle{splncs04}

\bibliography{miccaibib}
\end{document}